\documentclass[12pt,a4paper]{article}

\usepackage{graphicx}
\usepackage{xcolor}
\usepackage{hyperref}
\usepackage{subfig}
\usepackage{amsmath}
\usepackage[utf8]{inputenc}
\usepackage[T1]{fontenc}
\usepackage{authblk}
\usepackage{natbib}

\providecommand{\keywords}[1]{\textbf{\textit{Keywords---}} #1}

\title{Impact of Energy Efficiency on the Morphology and Behaviour of Evolved Robots}	
	
\author[1]{Margarita Rebolledo}
\author[2]{Daan Zeeuwe}
\author[1]{Thomas Bartz-Beielstein}
\author[2]{A.E. Eiben}
\affil[1]{Institute for Data Science, Engineering, and Analytics, TH Köln, Gummersbach, Germany}
\affil[1]{Department of Computer Science, Vrije Universiteit, Amsterdam, Netherlands}

\date{}	

\begin{document}

\maketitle
	
\begin{abstract}
Most evolutionary robotics studies focus on evolving some targeted behavior without taking the energy usage into account. This limits the practical value of such systems because energy efficiency is an important property for real-world autonomous robots. In this paper, we mitigate this problem by extending our simulator with a battery model and taking energy consumption into account during fitness evaluations. Using this system we investigate how energy awareness affects the evolution of robots. Since our system is to evolve morphologies as well as controllers, the main research question is twofold: (i) what is the impact on the morphologies of the evolved robots, and (ii) what is the impact on the behavior of the evolved robots if energy consumption is included in the fitness evaluation? The results show that including the energy consumption in the fitness in a multi-objective fashion (by NSGA-II) reduces the average size of robot bodies while at the same time reducing their speed. However, robots generated without size reduction can achieve speeds comparable to robots from the baseline set.
\end{abstract}


\section{Introduction}
Most evolutionary robot systems focus on some desired behavior of the robots taking the availability of energy for granted. In other words, the underlying models do not consider the battery and the fact that the battery charge is drained over time. This considerably increases the reality gap \citet{Jako95a, Mour17a} and limits the practical value of such systems.
	
The main motivation behind this paper is to make evolutionary approaches to robotics more relevant by fixing the `blind spot' on energy. In general, we propose that the traditional question
``What is the best robot controller that optimizes task performance for task X?'' be replaced by the question ``What is the best robot controller that optimizes task performance for task X and the energy efficiency of the robot?'' or by the question ``What is the best robot controller that optimizes task performance for task X and keeps the energy consumption under level Y?''
	
The matter of energy consumption is even more important in evolutionary robotics applications, where not only the controllers are being optimized by evolution, but also the morphologies. This is quite straightforward if we consider that in such a system the complete makeup of the robots can change, including the size, mass, and the number of components that use power, e.g., wheels or servo motors.
	
The study reported in this paper addresses the issue of energy efficiency in a system of morphologically evolving robots. The main objective is to gain insights into the effects of energy awareness on the evolutionary process. In particular, we research how the evolution of locomotion skills is affected if we include energy consumption into the notion of fitness. To be specific, we are interested in the result of the evolutionary process, that is the `dominant life forms' that evolve under these new conditions. This motivates the following research questions:  
	
\begin{description}
	\item[Q-1] How does the energy use affect the behavior of the evolved robots?
	\item[Q-2] How does the energy use affect the morphology of the evolved robots?
\end{description}
	
\noindent Regarding the evolved morphologies, we have an intuitive hypothesis:
\begin{description}
	\item[Hypothesis] Using energy efficiency as part of the robots' fitness will result in smaller robots with the same speed and/or faster robots with the same size.
\end{description}
	
To answer the above questions and to test the veracity of the hypothesis we extend an existing robot evolution system with the concept of energy. For this, a realistic model of energy consumption for each joint (servo motor) is included in the robots' body. 
To take into account the influence the energy information has on the robot's evolution we extend the system's fitness evaluation protocol, explained in more detail in subsequent sections.

In the current system, newborn robots are simulated for a given period of time and are evaluated by their speed, or equivalently, the distance they cover during the evalution period. 
This method can be extended naturally to evaluate the robots by their speed and the amount of energy they consume.
The resulting multi-objective optimization problem can be easily solved by either using a weighted average of the two objectives, or by a truly multi-objective method.

Reformulating the multi-objective problem as a single-objective problem via a weighted average as $F(x)= w_1 f_1(x) + w_2 f_2(x)$, requires the definition of the weights $w_1$ and $w_2$ by the user or by an automated tuning procedure. 
In the present study we choose the NSGA-II-based approach in order to reduce the effect of the user-defined weights or the increase of function evaluations required to properly tune the weight values. 

The structure of this work will be the following: in Section \ref{sec:related} we give a short overview of related works in the area of robot evolution with energy efficiency. Section \ref{sec:description} describes the system used for the experiments and the new battery module. In Section \ref{sec:experiments} we explain the experimental approach. Results are presented in Section \ref{sec:results}. Final thoughts and answers to our research questions are given in Section \ref{sec:conclusion}

\section{Related Work} \label{sec:related}
Robot locomotion is a common task in evolutionary robotics. In \citet{Kami03a} the generation of locomotion patterns in arbitrarily generated modular robots is studied. 
\citet{Lan18a} focuses on optimizing directed locomotion in evolvable robots. \citet{Haas10a} demonstrates how evolutionary algorithms can develop controllers for locomotion in aggregated organisms. Many other works focused on locomotion optimization can be found in the literature \citet{Marb05a, Oliv11a, Duar18a}\\

The topic of energy efficiency has been studied for robot components or robots with fixed morphologies. 
\citet{Gonz14a} compares the consumed energy and achieved speed between two types of robot legs.
\citet{Sapu15a} optimizes the energy efficiency of humanoid robots while maintaining good gait stability using a genetic algorithm.
\citet{Liu12a} optimizes the controller of biped robots to minimize the energy cost of locomotion.
\citet{Bing20a} uses reinforcement learning to optimize the controller of snake-like robots to generate energy efficient gaits.\\

Specifically related to the present work is the problem of energy consumption as part of the evolutionary process to obtain efficient robot morphologies and controllers able to locomote.
In \citet{Marg07a} multi-objective artificial evolution is used to optimize morphology and kinematic features for unmanned aerial vehicles.
\citet{Endo02a} optimizes the morphology and controller of humanoids robots to generate energy efficient walking robots.
To the best of our knowledge, there are not yet many works focusing on energy efficient evolvable robots.

\section{System Description} \label{sec:description}

\subsection{Simulator} \label{sub:simulator}
The robots used in this work are based on the framework RoboGen\footnote{\url{http://robogen.org}}.
In order to perform the simulations needed, the \textbf{R}obot \textbf{Evolve}\footnote{https://github.com/ci-group/revolve} (Revolve) \citet{Hup18a} toolkit was employed.
Revolve works on top of Gazebo\footnote{\url{https://gazebosim.org/}} and incorporates a set of tools to allow an easy definition of the robots, environments to execute the simulations, and objective functions to evaluate a robot's performance.

The robot's bodies are modular and composed of three modules: Core, Bricks and Joints.
The core component houses the robot's microcontroller and battery unit. Every robot can have only one Core.
The Bricks modules act as skeletal components and allow the attachmen of other components to any of its six faces. However, only flat morphologies are permited in our implementation, i.e., modules can only be attached to lateral faces of the Bricks and not the top or bottom.
The Joint module is an active element. It is the only component that provides the robot with the ability to move. Each Joint is powered by a servo motor and thus increases the energy consumption of the robot. 

The robot controller (``brain'') is defined as a neural network where every joint is mapped as an oscillator neuron.
Both morphology and controller for each robot are encoded by a Lindenmayer-System \citet{Jaco94a}, a parallel rewriting system and a formal grammar. A genotype is a tuple $G =(V,w,R)$ defining a grammar where $V$ is the alphabet, $w$ is the axiom and $R$ is the set of replacement rules. The starting point of a robot genotype is the axiom represented by the robots core module. The axiom is then iteratively extended into a longer sequence by executing the replacement rules $R$. The resulting string of symbols can then straightforwardly be mapped onto a morphology.

The definition of the evolutionary operators is included in the simulator. The crossover operation is performed by randomly taking production rules from the selected parents. The mutation is done by adding, deleting, or swapping one random element.
A more detailed information about the robots encoding and reproduction operators can be foun in \citet{Mira18a}

The simulator includes the definition of different terrains. For the present work only a flat environment without obstacles is used.  
For a simulation run, a robot is placed on the defined environment for a fixed amount of simulation steps. During the simulation the robot's performance according to a predefined fitness function is evaluated.

\subsection{Energy Consumption}\label{sub:battery}
A new battery module able to track the energy consumption of each robot was developed and added to Revolve to measure the energy consumption. 
As a starting point, we assume that the only components consuming power are the servo motors located at each of the Joints in the robot's body. The energy needed by the microcontroller, as well as, the idle energy consumption are neglected for the time being. 

The battery starts with an initial charge defined by the user and begins to discharge during the simulation time when the robot tries to move around the world. 
The amount of energy needed by an active servo motor is defined as its instantaneous power calculated as torque ($M$) times angular velocity ($\Phi$), given in $N \cdot m/s$. 
Following this, a robot's total energy consumption in one simulation step can be estimated as the sum of all its joints power as shown in equation \ref{eq:instapower}.
\begin{equation}
	\label{eq:instapower}
	\Delta C_i = \sum_{j=0}^{m}{ max( 0, M_{ij} \cdot \Phi_{ij}) }
\end{equation}
where $\Delta C_i$ is the robot's energy consumption at simulation step $i$, $m$ is the number of joints present in the robot's body and $M_{ij}$ and $\Phi$ are the torque and angular speed respectively.

The total energy consumption can be then determined as the sum of $\Delta C$ across all simulation steps.
In a realistic setting, a robot will not be able to continue functioning once the battery charge reaches zero nor would it make sense for the charge to reach negative levels. To avoid these situations in our simulation the constraint $E_{total}$ is defined as seen in \ref{eq_totalpower}.
\begin{equation}
	\label{eq_totalpower}
	E_{Total}= C_{Start} - \sum_{i=0}^{n}{\Delta C_i}
\end{equation}
where $C_{start}$ is the initial battery charge defined by the user, and $n$ is the number of simulation steps.\\
Once $E_{total}$ reaches zero the simulation is stopped independently of whether the defined simulation steps were achieved or not.
This means that robots with a higher energy consumption will have less time on the simulator than robots with better energy management.

\subsection{Speed}
The robot's speed is evaluated as the Euclidean distance between the robot's core position at the beginning and end of the simulation divided by the simulation time. It penalizes robots that do not move in a straight line. The speed is given in $cm/s$

\section{experiments} \label{sec:experiments}
Two different experiments are run to answer our research questions. The code used to run the simulations described in this paper can be found in \url{https://github.com/ci-group/revolve/tree/battery-master}

The first experiment will be our baseline to which we can compare possible differences in behavior or morphology caused by the battery inclusion. The baseline consists of an evolutionary run where speed is the only considered factor. A robots' objective is to evolve the best body-brain combination to achieve higher speeds.  

In the second experiment, the battery experiment, speed and energy consumption are simultaneously considered as objectives.
The evolutionary process will try to find robots able to move fast while saving as much battery as possible.
To estimate the best set of solutions for this multi-objective optimization problem the non-dominated sorting genetic algorithm 2 (NSGA-II) \citet{Deb02a} is implemented. The first objective is the robot's speed, the second objective is defined as the remaining battery after a simulation run, computed as explained in section \ref{sub:battery}. The goal is to maximize both objectives.

The size of the robots was limited to not allow more than 10 joints and not have more than 20 bricks.  
To determine the initial battery charge, $C_{Start}$, several simulations with robots of different sizes were run. It was observed that the average battery use after a simulation run was around $ 12 N m/s$ for a robot with the maximum size. To add pressure to the evolution the initial battery charge was set to $10 N m/s$.

For both experiments, the same evolutionary operators and parameters were applied. The population size was $\mu = 100$ evolved for 100 generations. In each generation parents were selected using tournament selection with size $k = 4$ and offspring $\lambda = 100$ were created. For the next generation, 100 individuals were selected again using tournament selection. 
The crossover and mutation operators were performed as explained in section \ref{sec:description} with a probability of 80\% each.
Each experiment was repeated 10 times to achieve statistical significance. During each repetition the system tries to learn the best body-controller combination that gives the best fitness evaluation.
Along with fitness measures we record for each robot other behavioral markers, such as the stability of the developed gait. Additionally, some mophological descriptos are also recorded such as the size of the robot, and the shape of the evolved morphology. A more in-depth review about the stored descriptors can be foun in \citet{Mira18a}.

\section{Results} \label{sec:results}
\begin{figure}
	\centering
	\includegraphics[width=0.9\linewidth]{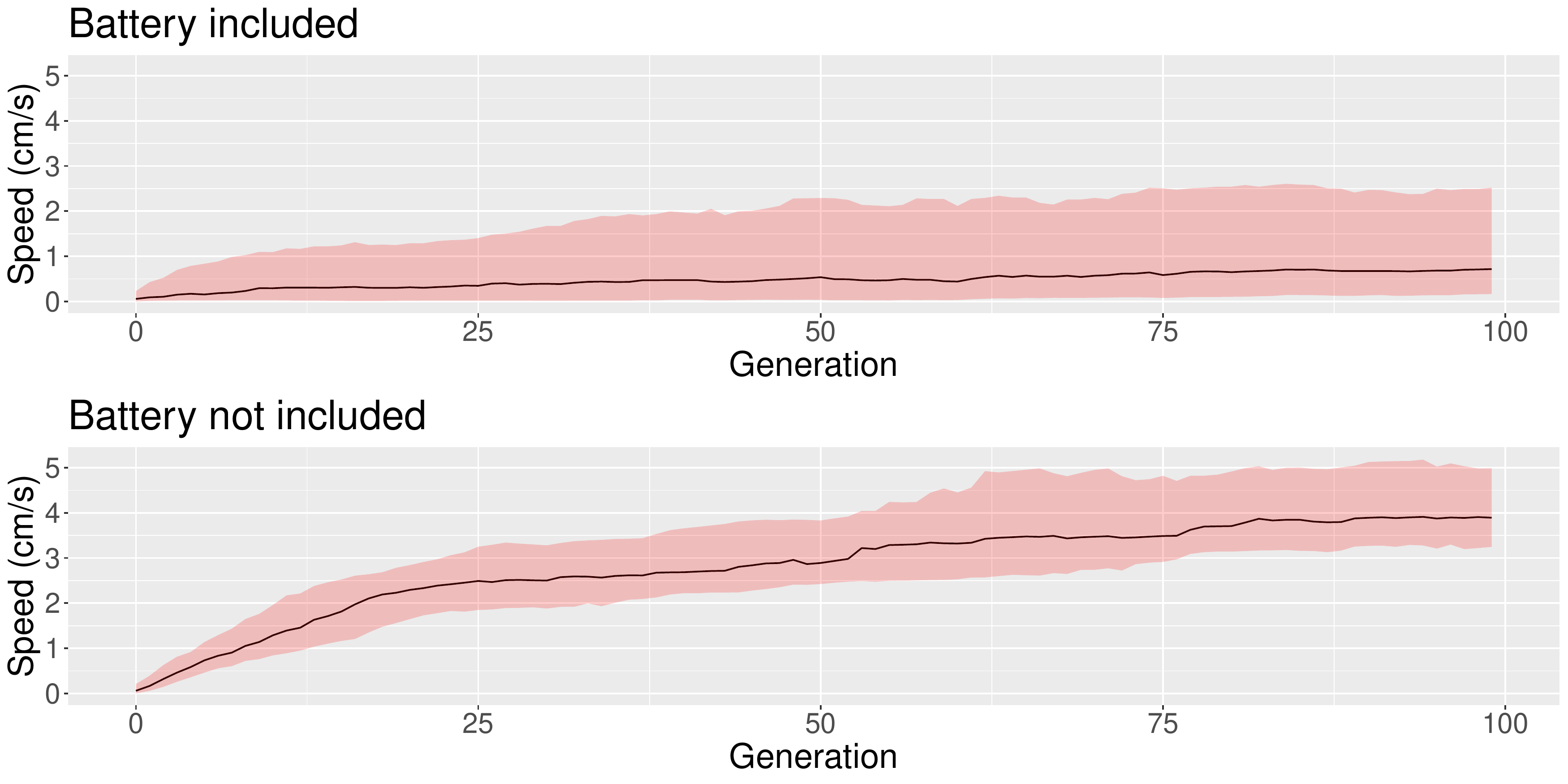}
	\caption{Speed comparison for the experiment with and without battery inclusion as a fitness measure. The solid line represents the median value of all robots in one generation across all ten repetitions. The first and third quartiles are also shown.}
	\label{fig:speed}
\end{figure}

\begin{figure}
	\centering
	\includegraphics[width=0.9\linewidth]{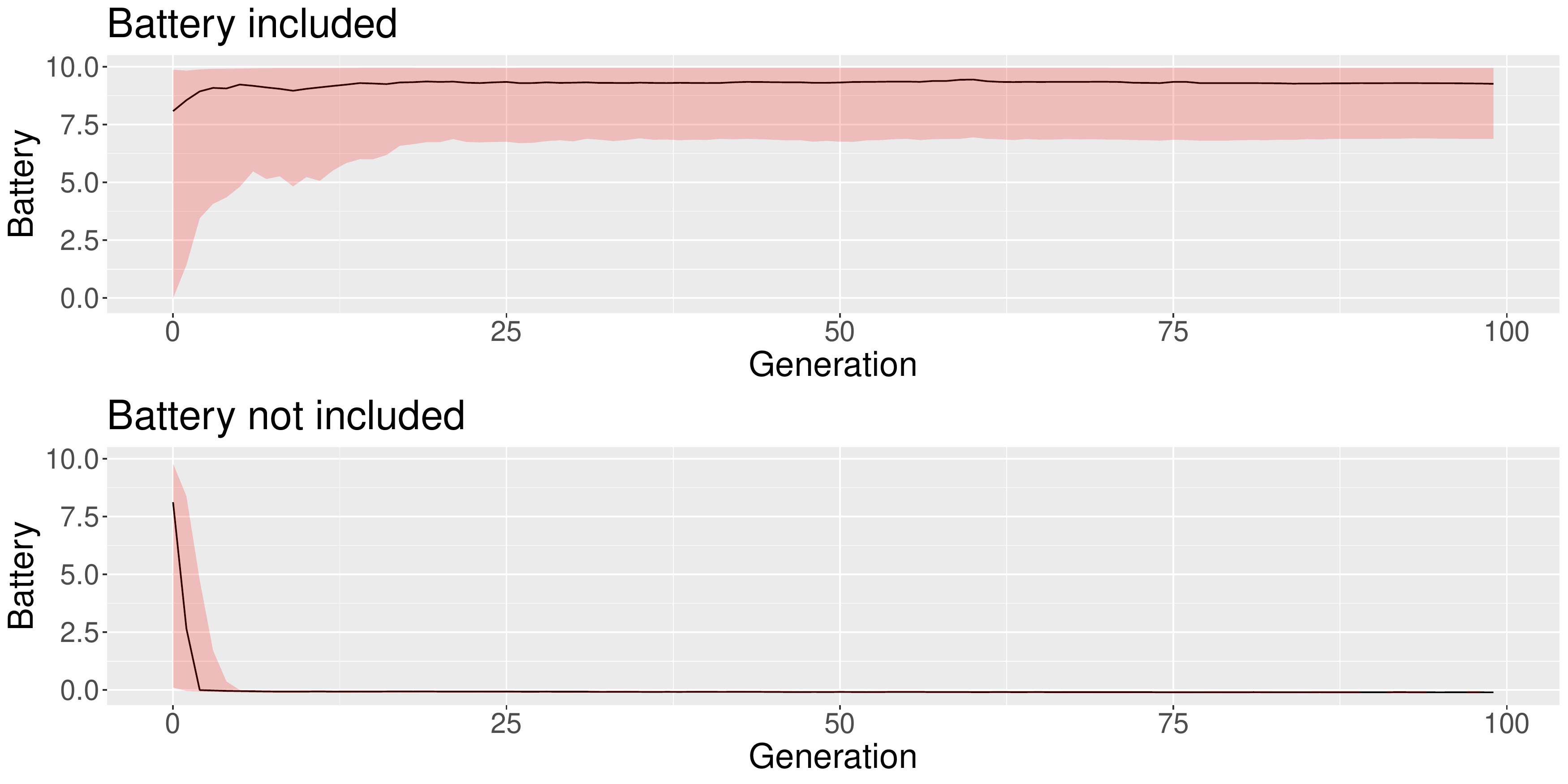}
	\caption{Remaining battery charge comparison for the experiments with and without battery inclusion as a fitness measure. The solid line represents the median value of all robots in one generation across all ten repetitions. The first and third quartiles are also shown. A value of 10 indicates no battery was used during the simulation, in this extreme case robots do not move during the simulation.}
	\label{fig:battery}
\end{figure}

As a starting point, we focus on the robot behavior during the evolutionary run.
Figures \ref{fig:speed} and \ref{fig:battery} showcase how speed and battery consumption change with respect to the generation respectively. The figures present the median value of speed or battery across the 10 simulation repetitions of the 1000 robots at each generation. The first and third quartile are also included to vizualise the stability of the results.  
The behavior differences between the two experiments are clear.
As seen in Figure \ref{fig:speed}, when compared to the baseline experiments, the median speed of the robots including energy consumption as a second fitness measure is considerably lower. The spread as indicated by the quartiles is also higher.
This outcome can be attributed to the normal behavior of the optimization algorithm. NSGA-II tries to find at each generation the set of nondominated solutions, in which each solution represent a trade-off between battery saver (no movement) and speedrunner (no battery saving). The inclusion of the extreme points at each generation drives the increase of the spread as shown in Figure \ref{fig:speed}

The remaining battery charge after a simulation run is shown in Figure \ref{fig:battery}. 
A value of 10 means no battery was used during the simulation. 
Robots not including energy fitness evaluation evolve very early to use all of the available battery charge. The algorithm learns that the addition of more joints or including more movement during the simulation is beneficial to achieve a better speed.


Another measurable behavior corresponds to the developed gait balance. It gives an insight into how robots move.
The balance is defined as the vertical angle the core module has in relationship to the ground during the simulation. It can range from 0 to 1. A value larger than 0.9 represents a very stable gait. Lower values could indicate that robots use rolling in their axis as locomotion method.
Observing the balance will allow us to know if there are any changes in the robot's evolved gait patterns.

\begin{figure}
	\centering
	\includegraphics[width=0.9\linewidth]{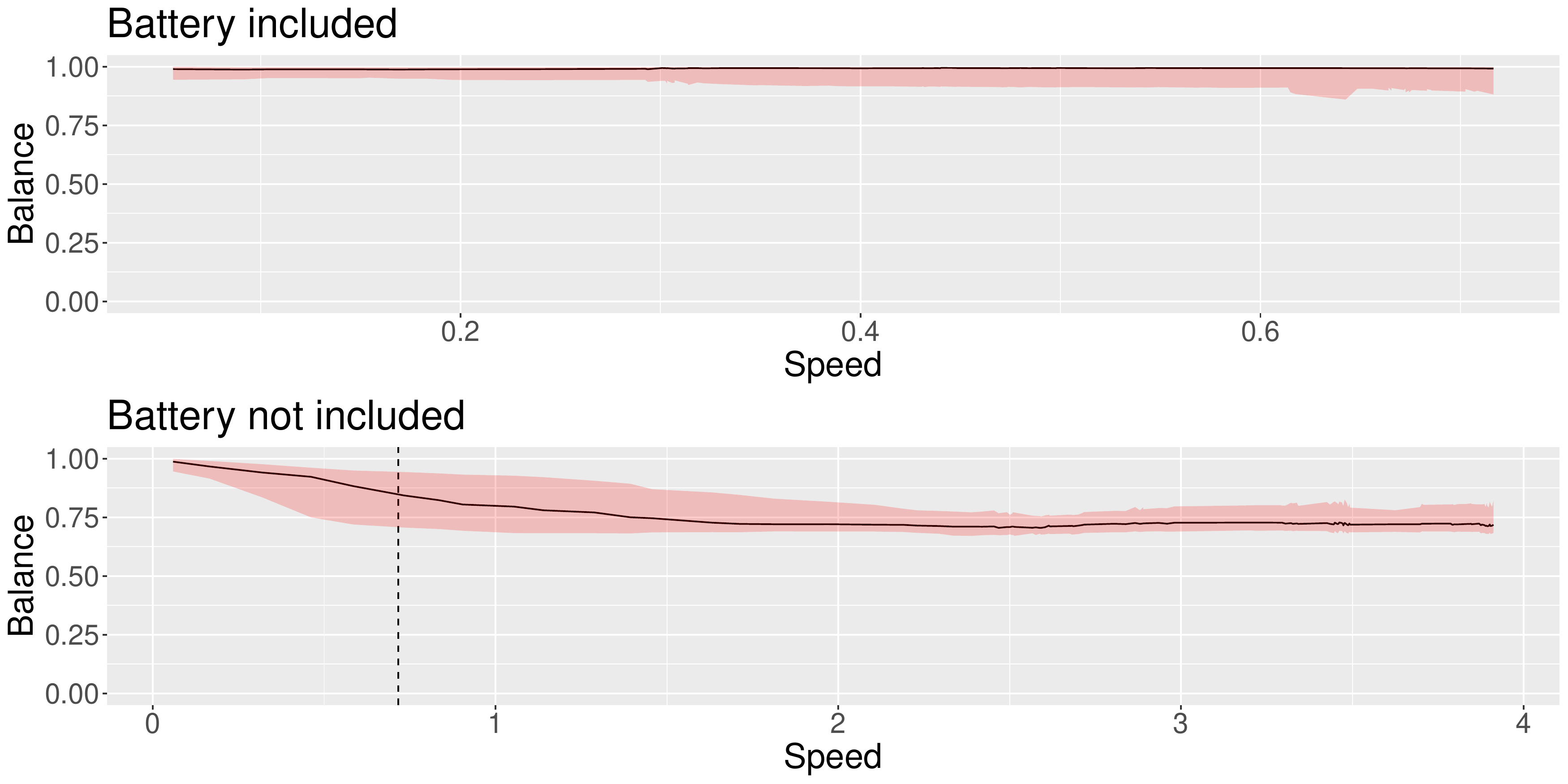}
	\caption{Balance comparison for the experiment with and without battery inclusion  as a fitness measure. The solid line represents the median value of all robots in one generation across all ten repetitions. The first and third quartiles are also shown. The vertical line indicates the maximum median speed value achieved in the battery experiment.}
	\label{fig:balance}
\end{figure}
Figure \ref{fig:balance} illustrates the balance behaviour for both experiments. Presented are the median of balance and speed across generations and runs.
Given the considerable difference between the median achieved speed for both experiments, only a portion of the balance for the battery experiment could be compared to the baseline. 
The dashed vertical line in Figure \ref{fig:balance}b) represents the maximum speed until which the balance can be compared in both experiments.
Robots including the battery constraint seem to present a more stable gait compared to robots with similar fitness in the baseline. 
To compare the balance for the faster robots the subset of robots within the battery experiment with a speed highter than 3.95$cm/s$ are considered. That is a speed higher than the median value.
For this subset of faster robots, the balance measure between both experiments is around 0.7. This implies that the energy usage does not greatly alter the gait behavior of faster robots.

\begin{figure}
	\centering
	\subfloat[Battery included]{\includegraphics[width=\linewidth,page=1]{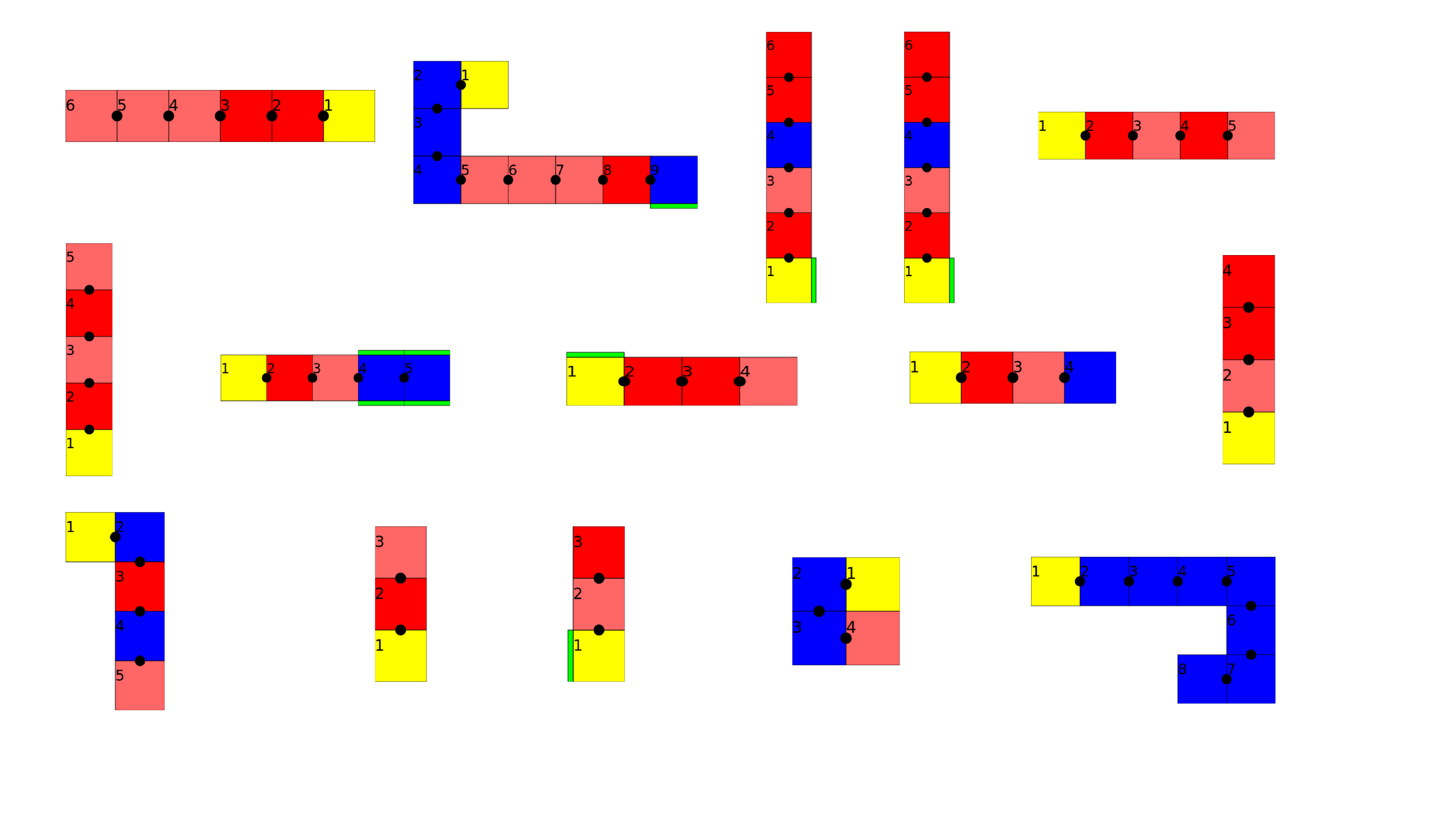}}\\
	\subfloat[Battery not included]{\includegraphics[width=\linewidth,page=2]{Figures/Bodies.pdf}}
	\caption{Robot morphologies comparison for evolution runs with and without battery. Selected are the robots composing the estimated Pareto front for Battery and Speed for generation 100 across all repetition runs. The color code indicates the module type, core (yellow), bricks (blue), or joint (red). Robots are ordered according to their achieved speed from left (higher) to right (lower). The robot in the bottom right corner is the slowest robot.}
	\label{fig:bodies}
\end{figure}

To answer our second research question we focus on the robot's morphology. We present a sample of the nondominated resulting robot morphologies for both experiments in Figure \ref{fig:bodies}. 
They robot bodies presented are ordered from left to right in decreasing speed.
For the battery experiment it can be observed that highly battery efficient robots, bottom right corner, are included in the form of only brick bodies unable to move. These robots save the most battery while staying the maximum permitted time in the simulator. 

\begin{figure}
		\centering
	\includegraphics[width=0.8\linewidth,page=1]{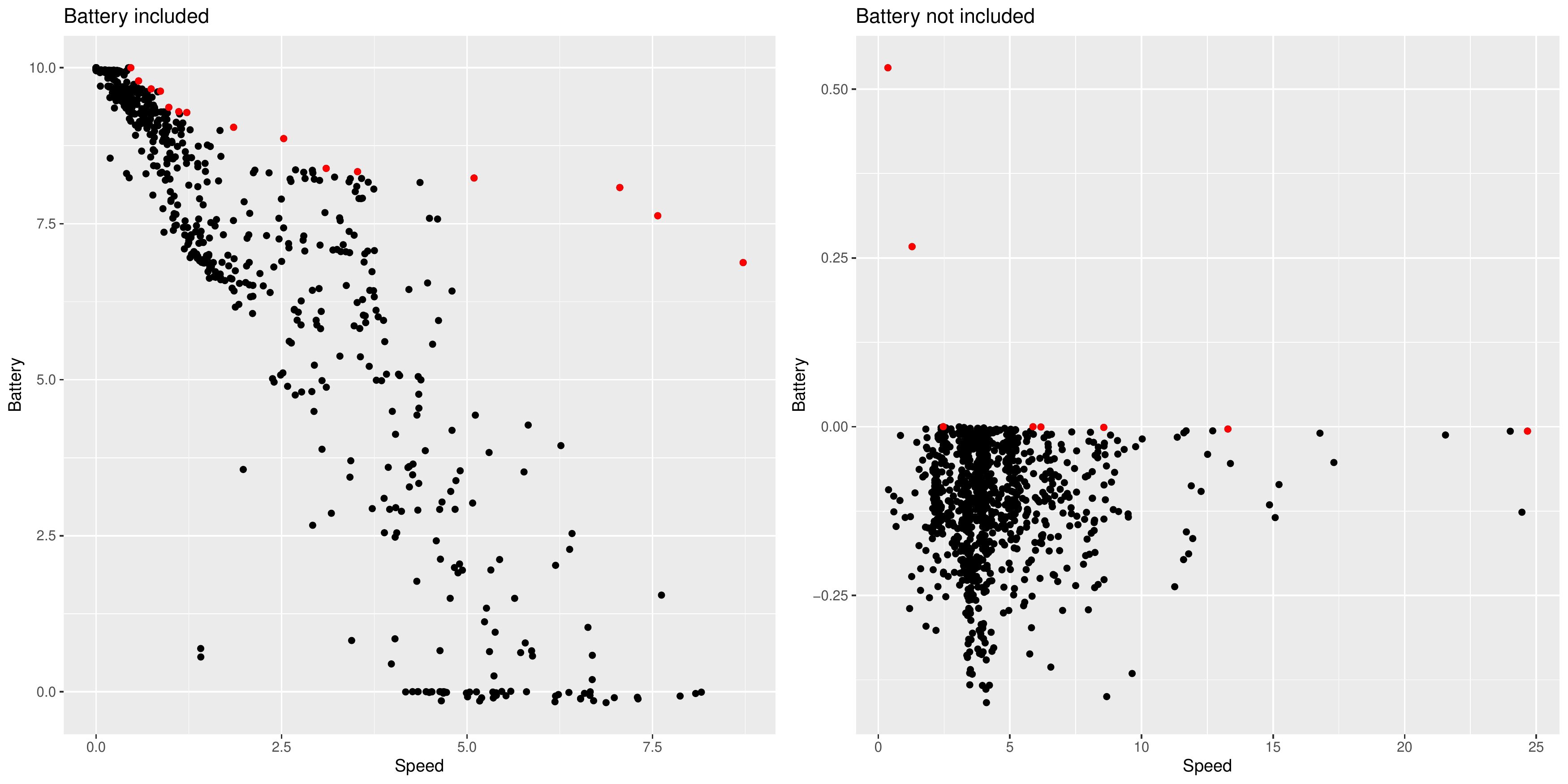}
	\caption{Set of robots from the last generation of all 10 repetitions. Each point represents a robot ordered by its battery consumption and speed. The nondominated points are illustrated in red.}
	\label{fig:pareto}
\end{figure}

To select the morphologies in Figure \ref{fig:bodies} all the robots in the last generation of all 10 repetitions were considered. Figure \ref{fig:pareto} shows these robots in terms of their battery consumption and speed. The nondominated points are illustrated in red. 

Morphologies in both experiments converge mostly to snake-like forms. This form has been shown to be one of the most effective bodies for locomotion in the implemented evolution and encoding system \citet{Jeli18a}.
One of the most noticeable differences between the resulting morphologies is the number of joints included in the body.This is a good evidence that the robot is trying to save energy.Robots evolved with the battery constraint can achieve a similar speed as robots from the baseline while having a reduced size.
A larger body is an advantage in terms of achieving the largest possible displacement. However, observing the resulting morphologies and corresponding speeds, it can be said that smaller robots can also evolve good locomotion skills. This statement aligns with one of our initial hypothesis.
A visual inspection of the gaits evolved by the nondominated morphologies was conducted. The differences in movement are not considerable. The faster robots in both experiments tend to move by rotating in their own axis.

The speed and size of the resulting in generation 100 for all repetitions are compared. The objective is to gain more insight into the size/speed behavior of the robots with battery constraint.
We define a minimum speed threshold of $7  cm/s$ as an acceptable speed for a robot to achieve.

\begin{figure}
	\centering
	\includegraphics[width=0.8\linewidth]{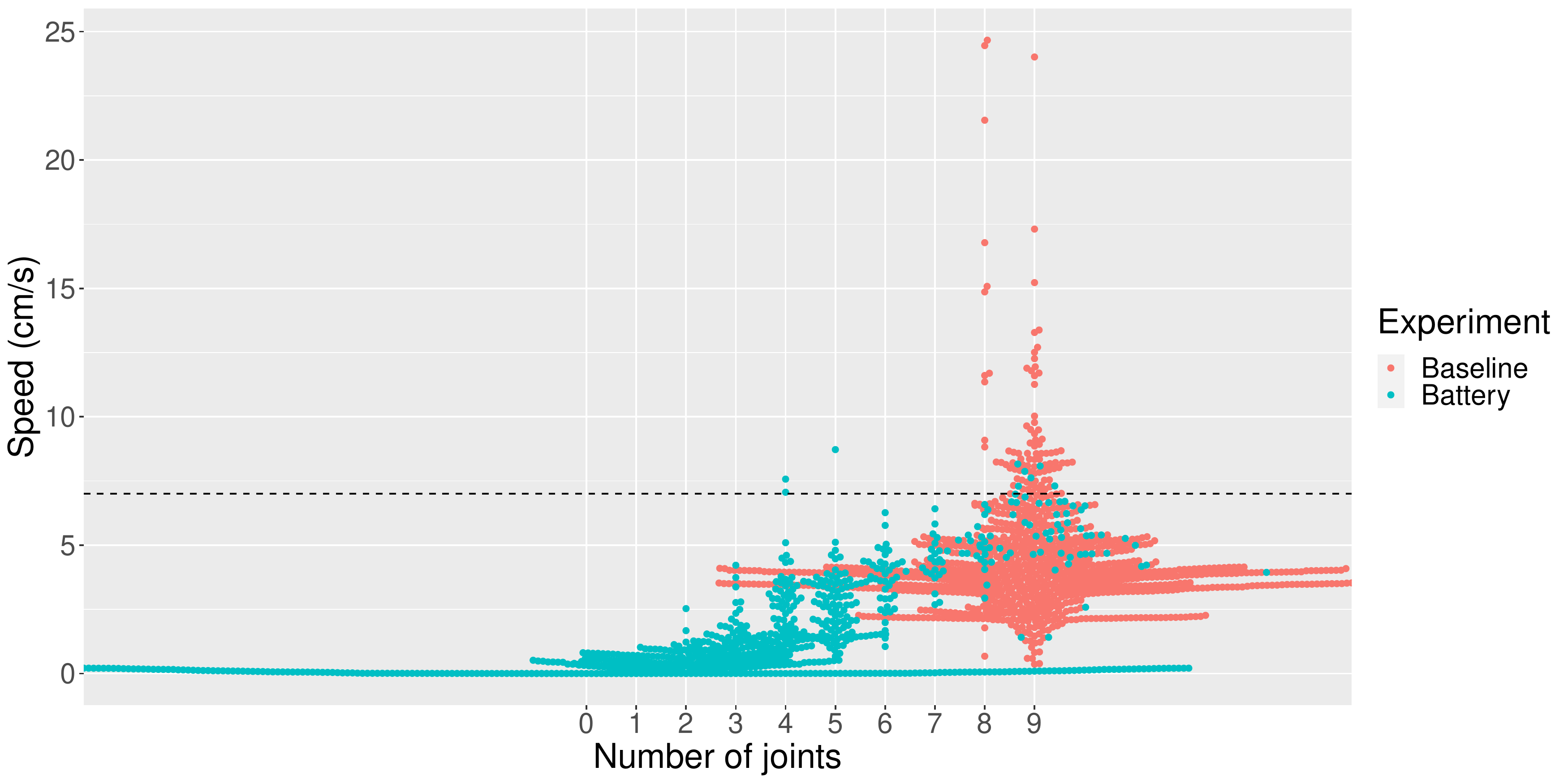}
	\caption{Robot distribution around speed values given their number of joints. Illustrated are the resulting robots in generation 100 for all repetition runs}
	\label{fig:size_speed}
\end{figure}
Figure \ref{fig:size_speed} illustrates distribution of the last generation robots for both experiments. The robots are presented according to their size and speed. The dashed line ilustrates the defined $7  cm/s$ threshold.

If we focus only on the robots that achieved a speed higher than the threshold it is seen that only a small number of robots from the battery experiment are above it. 
Recalling our initial hypothesis, we can say that the experiments do not support the first option stating that the inclusion of battery will result in smaller robots with the same speed.
A summary of the robots located above the $7 cm/s$ threshold is given in Table \ref{tab:speedThres}. A Welch's t-test (p>0.05) indicates that the difference in number of joints between bot experiments is not significant.

\begin{table}[]
	\centering
	\caption{Mean number of joints for robots with a speed of over $7 cm/s$}
	\label{tab:speedThres}
	\begin{tabular}{lccc}
		\hline
		Experiment & No. Robots & Mean Number Joints & SD\\
		\hline
		Battery & 9 & 7.44 & 2.35 \\
		Baseline & 91 & 8.88 & 0.32\\
		\hline
	\end{tabular}
\end{table}

This data raises another interesting question on the mean speed achieved by robots of similar sizes. For this, we take the same group of resulting robots, but focus only on robots with 9 joints. 
Table \ref{tab:sizeThres} summarises the results for both experiments. A Welch's t-test (p<0.001) indicates the difference between the two groups is significant.

\begin{table}[]
	\centering
	\caption{Mean speed for both experiments for the case of robot with over 8 Joints}
	\label{tab:sizeThres}
	\begin{tabular}{lccc}
		\hline
		Experiment & No. Robots & Mean speed & SD\\
		\hline
		Battery & 71 & 5.35 & 1.29 \\
		Baseline & 940 & 4.33 & 2\\
		\hline
	\end{tabular}
\end{table}
The results seem to corroborate the second part of our initial hypothesis, the battery constrained experiment will result in robots of the same size, but faster on average than the baseline. A larger number of simulation steps could be the reason why the energy efficient robots were able to achieve a better speed. 
This kind of behavior could be especially useful if the robots need to undergo an additional learning phase to improve their controller in a restricted simulation time.

\section{Concluding remarks} \label{sec:conclusion}
This paper addresses an issue that has a great impact on the real-world applications of evolutionary robotics, yet has been largely ignored in the past. Our main objective is to gain insights into the effects of `energy awareness' on robot evolution. To be specific, we are interested in the result of the evolutionary process, that is, the behavior and the morphology of the evolved robots if moving around is not free, but has energy costs. 

To achieve this objective, we have extended our current system to model energy consumption. The consumed energy is calculated as the instantaneous power of the servo motors in the robot's body. For this study the energy consumed by the microcontroller and the idle energy consumption are neglected. This is not a severe limitation, however, because in our system\footnote{We mean our physical robot system that is the basis of the simulated one. Actually, we have a dual system, where each robot we can simulate can also be constructed in reality and vice versa.} moving needs more energy than thinking, i.e., the servos consume much more energy than the processors.

In order to answer our research questions and assess our initial hypothesis two series of experiments were conducted. The first series of experiments comprises our baseline data and consists of an evolutionary run with speed as the single objective.
In our second series of experiments we define energy consumption and speed as the two objectives for robot evolution. To solve the multi-objective problem without a user-defined parameter (the relative weights of the two objectives when redefining the problem as a weighted average) we apply the NSGA-II algorithm. 

Based on the results of these experiments we can answer our research questions as follows:
\begin{description}
	\item[Q-1] {\it How does the energy use affect the behavior of the evolved robots?} \\
	We measure the behavior of the evolved robots as the speed achieved at the end of their simulation time. The median speed along the evolutionary run was considerably lower for the robots with battery limitations. This is exacerbated by the inclusion of robots with high battery efficiency but no ability to move as they are also part of the estimated Pareto front. This explains the higher spread seen on the speed data and does not exclude the possibility of obtaining robots with high speed. The balance behavior indicates that robots with battery limitations tend to move differently at lower speeds but this behavior disappears for higher speeds.

	\item[Q-2] {\it How does the energy use affect the morphology of the evolved robots?} \\ 
	The most visible influence on the morphology of the robots is the reduction in the number of joints along the evolutionary process. 
	Morphologies with the ability to move mostly converged to a snake-like shape which from the baseline and other experiments using the same system seems to be one of the most efficient shapes for locomotion in this specific system.
\end{description}

Further to these questions we have also formulated a hypothesis regarding the evolved morphologies: Using energy efficiency as part of the robots' fitness will result in smaller robots with the same speed and/or faster robots with the same size. The experimental results seem to support the hypothesis that robots of the same size can be faster when evolution takes energy consumption into account.

In summary, with this paper we attempt to put the issue of energy consumption on the research agenda for evolutionary robotics. We argue that this development will make the field practically more relevant. In the meanwhile, it will also increase the biological plausibility of more theoretically inclined studies. 

Our current work demonstrates `energy aware' evolutionary robotics systems and delivers the first insights into their workings. We are fully aware of the limitations, we considered only one task --albeit one of the most fundamental one-- and only one specific way of combining the two objectives of task performance and energy efficiency. The use of other multi-objective optimization algorithms as well as different tasks to define the fitness function are subject to further research.

\bibliographystyle{plainnat}
\bibliography{rebo21a}

\end{document}